\documentclass[runningheads]{llncs}
\usepackage{subcaption}
\usepackage{booktabs}
\usepackage{siunitx}
\usepackage[T1]{fontenc}
\usepackage{graphicx}
\usepackage{booktabs}
\usepackage{hyperref}
\usepackage{color}

\usepackage[utf8]{inputenc}
\usepackage{tikz}
\usepackage{pgfplots}
\usepgfplotslibrary{groupplots,dateplot}
\usetikzlibrary{patterns,shapes.arrows}
\pgfplotsset{compat=newest}

\usepackage{amssymb}
\usepackage{multirow}
\usepackage{booktabs}
\usepackage{siunitx}
\usepackage{colortbl,xcolor}

\usepackage{array}
\newcolumntype{C}{>{\centering\arraybackslash}p{1.1cm}}
\newcolumntype{D}{>{\centering\arraybackslash}p{1.65cm}}

\usepackage{orcidlink}

\begin{document}

\title{ClapperText: A Benchmark for Text Recognition in Low-Resource Archival Documents}
\titlerunning{ClapperText}
%
\author{Tingyu Lin\orcidlink{0009-0008-9825-686X} \and
Marco Peer\orcidlink{0000-0001-6843-0830} \and
Florian Kleber\orcidlink{0000-0001-8351-5066} \and
Robert Sablatnig\orcidlink{0000-0003-4195-1593}}

\authorrunning{T. Lin et al.}

%
\institute{Computer Vision Lab, TU Wien\\
\email{tylin@cvl.tuwien.ac.at}}
\maketitle              
\begin{abstract}

This paper presents ClapperText, a benchmark dataset for handwritten and printed text recognition in visually degraded and low-resource settings. The dataset is derived from 127 World War II-era archival video segments containing clapperboards that record structured production metadata such as date, location, and camera-operator identity. ClapperText includes 9,813 annotated frames and 94,573 word-level text instances, 67\% of which are handwritten and 1,566 are partially occluded. Each instance includes transcription, semantic category, text type, and occlusion status, with annotations available as rotated bounding boxes represented as 4-point polygons to support spatially precise OCR applications. Recognizing clapperboard text poses significant challenges, including motion blur, handwriting variation, exposure fluctuations, and cluttered backgrounds, mirroring broader challenges in historical document analysis where structured content appears in degraded, non-standard forms. We provide both full-frame annotations and cropped word images to support downstream tasks. Using a consistent per-video evaluation protocol, we benchmark six representative recognition and seven detection models under zero-shot and fine-tuned conditions. Despite the small training set (18 videos), fine-tuning leads to substantial performance gains, highlighting ClapperText’s suitability for few-shot learning scenarios. The dataset offers a realistic and culturally grounded resource for advancing robust OCR and document understanding in low-resource archival contexts. The dataset and evaluation code are available at \url{https://github.com/linty5/ClapperText}.

\keywords{Low-Resource Document Analysis \and Handwritten Text Recognition \and Archival OCR \and Scene Text Benchmark}

\end{abstract}
\section{Introduction}

\begin{figure}[htb]
  \centering
  \begin{subfigure}[b]{0.95\textwidth}
    \includegraphics[width=\linewidth]{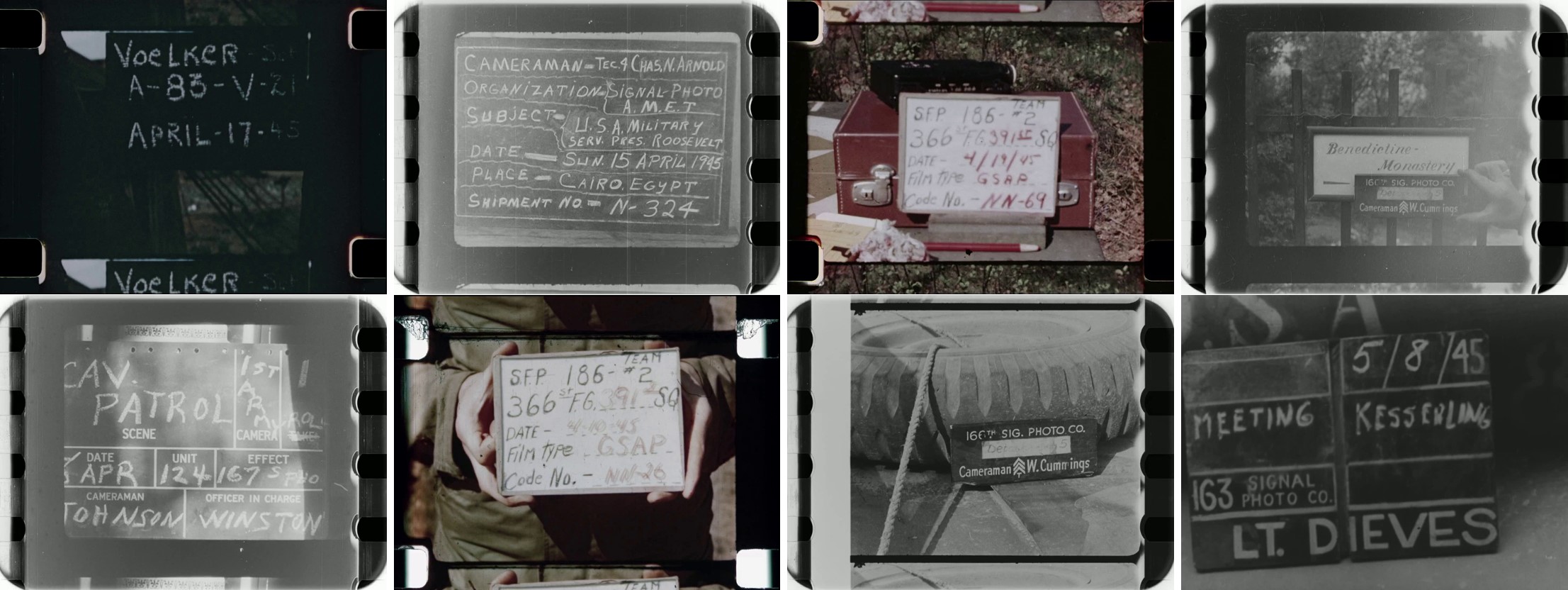}
    \caption{Samples from the training set.}
  \end{subfigure}
  \hfill
  \begin{subfigure}[b]{0.95\textwidth}
    \includegraphics[width=\linewidth]{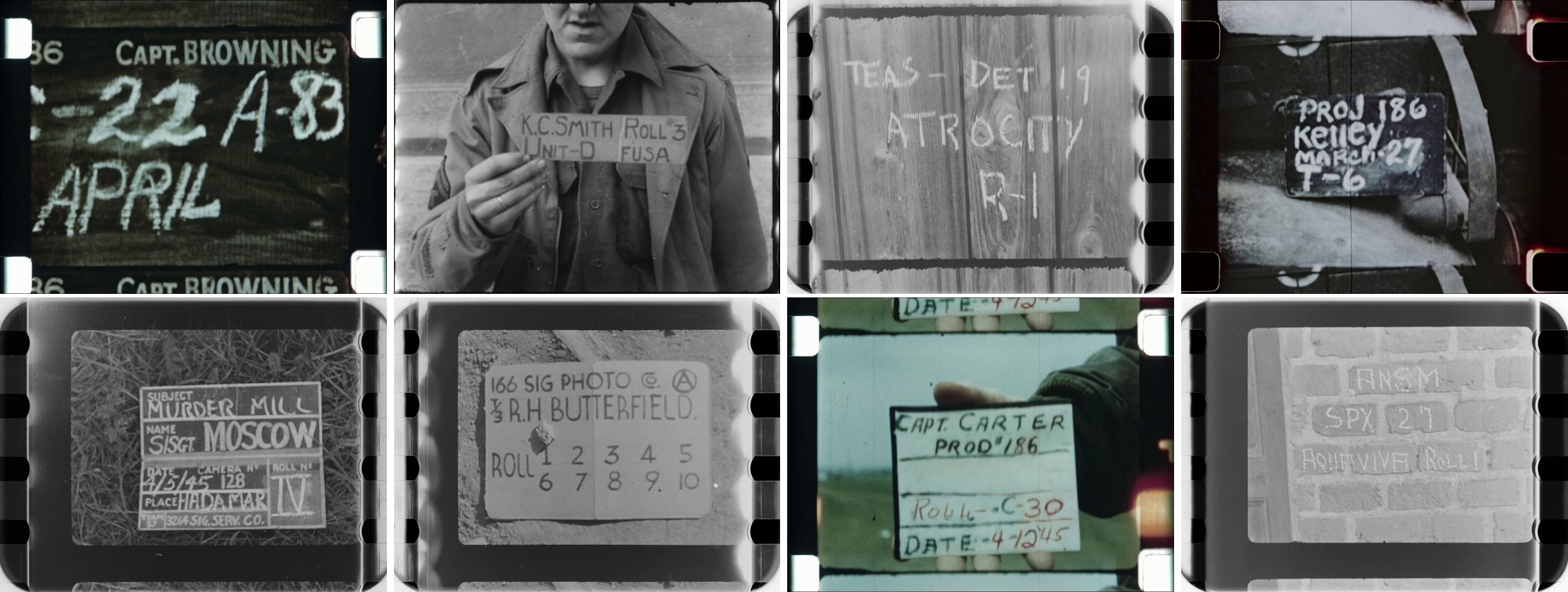}
    \caption{Samples from the test set.}
  \end{subfigure}
  \caption{Visual examples of annotated keyframes in ClapperText. Each subfigure shows eight frames, illustrating variation in layout, handwriting style, and image quality across splits.}
  \label{fig:split_examples}
\end{figure}

Text appearing in historical film footage provides valuable metadata for archiving, retrieval, and historical analysis. Clapperboards act as semi-structured visual records, capturing production details such as date, location, and camera-operator information crucial to historians and archivists. Figure~\ref{fig:split_examples} illustrates the visual diversity of historical clapperboards. While recent advances in optical character recognition (OCR) have markedly improved text detection and recognition in modern documents and natural scenes~\cite{chen2021text,long2021scene,wang2023survey}, OCR research targeting archival audiovisual materials, especially those degraded by motion blur, exposure shifts, and handwriting variation, remains limited.

Existing handwritten datasets cover modern and historical documents, but primarily in the form of high-resolution scanned pages. Modern collections~\cite{kleber2013cvl,marti2002iam,mondal2024bridging} offer stylistic diversity under clean visual conditions, while historical corpora~\cite{clerice2024catmus} often suffer from degradation such as fading or physical damage. However, neither reflects handwritten text's visual and structural inconsistencies in audiovisual archives. In parallel, scene text benchmarks like ICDAR 2015~\cite{karatzas2015icdar} and COCO-Text~\cite{veit2016coco} have advanced recognition in natural images. More recent video text datasets expand into real-world domains, but primarily focus on structured content such as subtitles~\cite{tian2017unified} or traffic signs~\cite{reddy2020roadtext}, with consistent layouts and clean annotations. These characteristics contrast sharply with the fragmented, handwritten, and visually degraded text found in historical clapperboards, which remains underexplored.

While our work focuses on OCR in historical film footage, it aligns with the goals of traditional low-resource language analysis through its emphasis on underexplored, culturally significant, and technologically challenging textual data. The clapperboards in archival films contain structured handwritten and printed metadata—such as dates and locations—that, though not belonging to a single language category, exist in a highly degraded and low-resource modality. These texts, embedded in deteriorated historical video, pose similar challenges to low-resource language document analysis, including data scarcity, handwriting variation, and poor visual conditions. As such, these materials represent a compelling instance of low-resource historical textual data that benefits from targeted document analysis techniques.

\begin{figure}[htb]
    \centering

\definecolor{myblue}{RGB}{20,112,176}

\begin{tikzpicture}
\begin{axis}[
    ybar,
    width=0.9\textwidth,
    height=0.3\textwidth,
    ylabel={Number of Videos},
    symbolic x coords={0--49,50--99,100--149,150--199,200--249,250+},
    xtick=data,
    bar width=15pt,
    enlarge x limits=0.1,
    ymin=0,
    ytick={0,10,20,30,40,50},
    ymajorgrids=true,
    grid style=dashed,
]
\addplot[fill=myblue] coordinates {
    (0--49,49)
    (50--99,47)
    (100--149,17)
    (150--199,8)
    (200--249,3)
    (250+,3)
};
\end{axis}
\end{tikzpicture}
    \caption{Distribution of frame counts per video segment in ClapperText.}
    \label{fig:frame_distribution}
\end{figure}
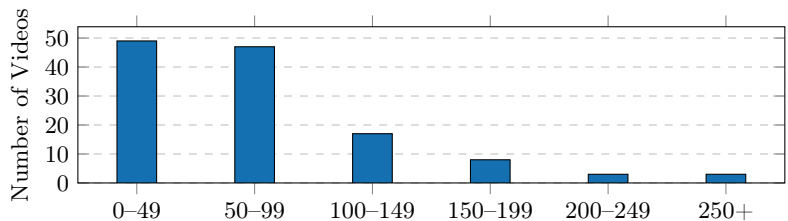

This work addresses these limitations by focusing on clapperboard frames from archival films. The HISTORIAN dataset~\cite{helm2022historian} provides a large-scale collection of digitized World War II-era film reels, accompanied by detailed annotations on shot boundaries, cinematographic styles, and camera movements. However, it lacks the textual annotations necessary for document-level analysis, such as word transcriptions, text categories, and layout information. To bridge this gap, we introduce \textbf{ClapperText}, a dataset of 127 curated video segments extracted from HISTORIAN, annotated at the word level with high spatial precision and semantic structure. As shown in Figure~\ref{fig:frame_distribution}, most segments fall within short to moderate duration ranges, providing a mix of concise and extended temporal sequences suitable for frame-level annotation.

ClapperText contains 9,813 fully annotated video frames and 94,573 labeled text instances, of which 1,566 are marked as partially occluded. The dataset exhibits substantial visual diversity: over 67\% of text instances are handwritten, and annotations span multiple categories including \texttt{Text}, \texttt{Date}, \texttt{Location}, \texttt{Recorded\_By}, and \texttt{Attribute}. This semantic structure aligns with historical clapperboard conventions and supports understanding of fine-grained documents. Unlike synthetic or clean datasets, ClapperText captures challenging real-world properties such as motion blur, camera shake, inconsistent exposure, and historical handwriting styles.

This paper introduces ClapperText and evaluates modern OCR systems in a low-resource, visually degraded setting. Our contributions are threefold:
\begin{itemize}
    \item \textbf{Dataset.} We release ClapperText, a frame-level dataset of 9,813 keyframes with 94,573 word-level annotations (handwritten, occluded, and category-labeled) extracted from 127 World War II film segments.
    \item \textbf{Benchmarks.} We benchmark six text-recognition and seven text-detection models under zero-shot and fine-tuned conditions, analysing how modern OCR adapts with limited supervision.
    \item \textbf{Analysis.} We provide detailed accuracy breakdowns by text type, highlighting the benefits and limitations of fine-tuning in few-shot historical settings.
\end{itemize}

By releasing the dataset and strong baselines, we aim to accelerate research on document understanding in culturally significant yet under-annotated archival material.

\section{Related Work}

Several foundational datasets have established benchmarks for scene text detection and recognition. The ICDAR 2003 \cite{1227749} and ICDAR 2013 \cite{6628859} datasets introduced text detection challenges under controlled conditions, primarily focusing on horizontally aligned text. Subsequent datasets addressed more complex scenarios. The ICDAR 2015 dataset \cite{karatzas2015icdar}, for instance, provided large-scale annotations for text appearing in real-world environments, including incidental and blurred text instances. The IIIT 5K-word dataset \cite{MishraBMVC12}, compiled from Google image searches, presents a diverse collection of cropped word images. The COCO-Text dataset \cite{veit2016coco} extends the variety of text appearances by leveraging the COCO object recognition dataset, offering over 63,000 images with text annotations. The Total-Text dataset \cite{ch2017total} introduced a benchmark for curved text detection, while the HOST dataset \cite{wang2021two} specifically focused on occluded text recognition. Additionally, synthetic datasets such as Syn90k \cite{Jaderberg14c,Jaderberg16} and SynthText \cite{Gupta16} have been widely used to generate large-scale training samples. Long et al.~\cite{long2022towards} propose a unified end-to-end framework that jointly performs scene text detection and layout analysis, introducing HierText, a densely annotated dataset with hierarchical text structures in natural scenes. Despite their diversity and scale, these image-based datasets do not accurately capture the degradation effects found in historical film archives.

Recent datasets have addressed the challenges of text recognition in videos, which involve dynamic motion and varying lighting conditions. The dataset introduced by Tian et al. \cite{tian2017unified} primarily targets subtitle and caption recognition in online videos, focusing on horizontally aligned text. The RoadText-1K dataset \cite{reddy2020roadtext} extends video-based text analysis to driving scenarios, emphasizing the detection of text on street signs and traffic signals. While these datasets accommodate real-time variations, they do not encompass the severe degradation in aged film footage. Furthermore, contemporary video-based scene text datasets rarely incorporate handwritten scripts or the combined effects of motion blur and archival deterioration, underscoring the need for specialized datasets tailored to historical film analysis.

\section{The ClapperText Dataset}

\begin{figure}[htb]
    \centering
    \includegraphics[width=\linewidth]{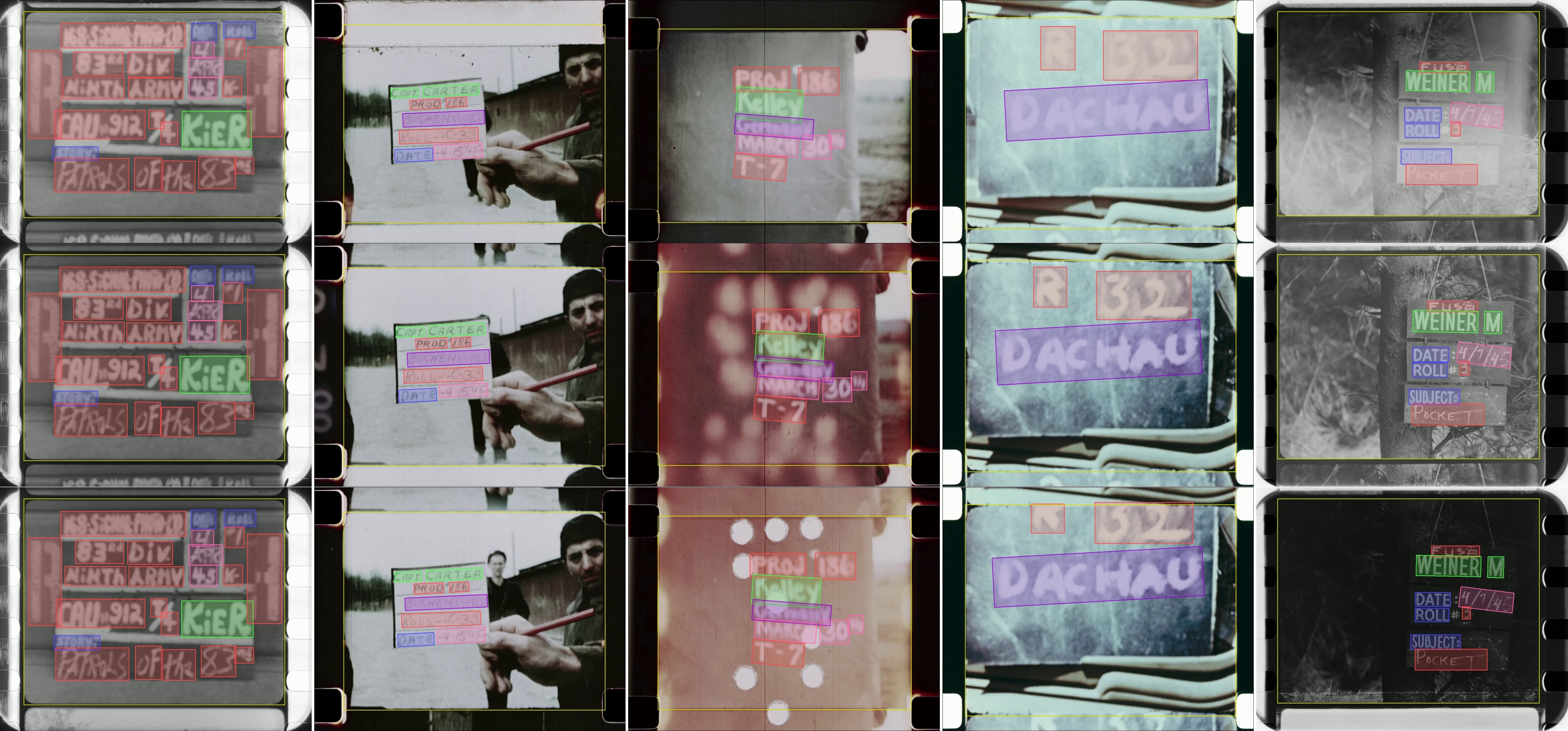}
    \caption{Keyframe samples from five ClapperText videos. Each column illustrates a distinct intra-video variation: (1) minimal change across dense frames, (2) background variation outside the clapperboard, (3) color shift and damage obscuring text, (4) clapperboard motion affecting visibility, and (5) brightness fluctuation.}
    \label{fig:frame_examples}
\end{figure}

We selected 127 clapperboard‐containing video segments from more than 300 candidates in HISTORIAN. All segments retain their original recording resolution of 1440 × 1080 pixels and a playback rate of 24 frames per second. Rather than pursuing complete coverage, we prioritised visual and semantic diversity. Many excluded segments showed near-duplicate clapperboard layouts, text styles, or backgrounds already represented in the selected set. This filtering allows ClapperText to capture a broad range of historical characteristics—such as handwriting variation, exposure instability, motion blur, and occlusion—while avoiding redundancy. Figure~\ref{fig:frame_examples} presents representative keyframes from five video segments, highlighting the intra-video variation that motivated our decision to annotate at the frame level. Even within a single segment, camera movement or lighting changes can alter text visibility and quality, making each frame a distinct OCR challenge.

\subsection{Text Annotation Process}\label{sec:annotation}

\begin{figure}[htb]
    \centering
    \includegraphics[width=\linewidth]{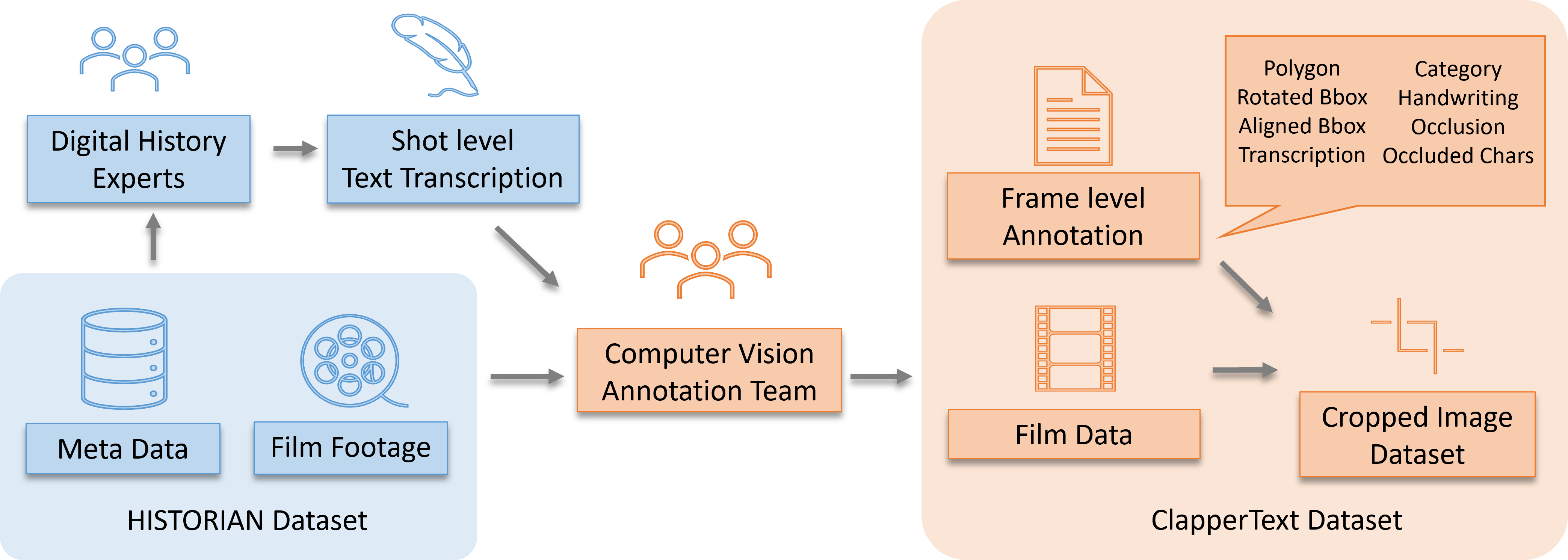}
    \caption{Annotation pipeline: historians transcribe shot-level text; computer vision annotators label text instances in video frames.}
    \label{fig:annotation_process}
\end{figure}

The ClapperText dataset was constructed through a multi-stage annotation workflow involving historical and technical experts. As shown in Figure~\ref{fig:annotation_process}, a team of historians first produced shot-level transcriptions of the clapperboard content based on available metadata and visual inspection. This ensured transcription fidelity and semantic accuracy aligned with historical records.

A computer vision team then annotated video frames using the Computer Vision Annotation Tool (CVAT) \cite{cvat_ai_corporation_2024_12771595}. The annotation followed a three-stage review process: an initial annotator created bounding boxes and metadata, which were refined by a second reviewer and audited by a third. Disagreements were resolved through consensus. All text regions were annotated at the word level to ensure consistency across frames and categories. To capture temporal variation with minimal redundancy, annotators labeled at least five keyframes per video, adding more for unstable sequences (e.g., motion blur, exposure shifts). Intermediate frames were interpolated via CVAT and manually verified. Keyframes were densely spaced, with a maximum gap of 12 frames (0.5s) and an average of 5.58 frames (0.23s), ensuring that short-lived visual changes were well represented. 

\begin{figure*}[htb]
  \centering
  \begin{subfigure}[b]{0.45\textwidth}
    \centering
    \includegraphics[width=\linewidth]{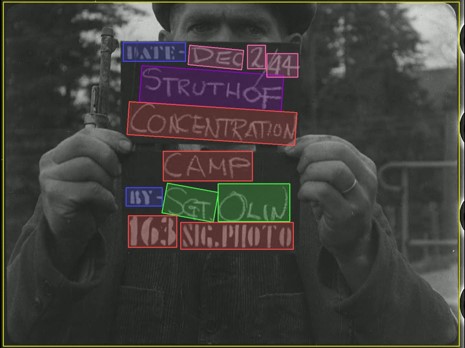}
    \caption{Frame annotations}
    \label{fig:anno_frame}
  \end{subfigure}
  \hfill
  \begin{subfigure}[b]{0.54\textwidth}
    \centering
    \includegraphics[width=\linewidth]{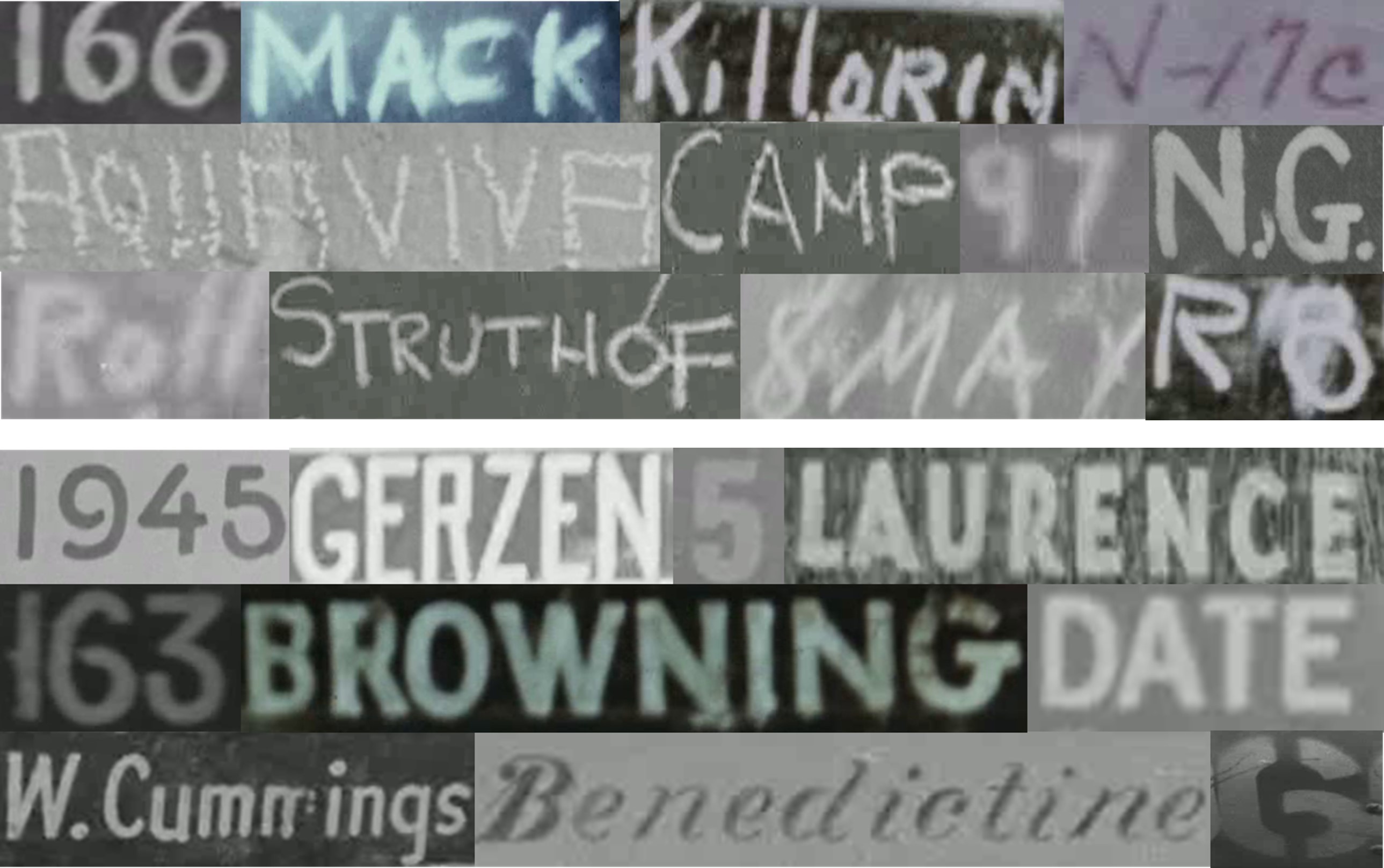}
    \caption{Cropped text images}
    \label{fig:crop_collage}
  \end{subfigure}
  \caption{Examples of annotated data in ClapperText. (a) shows word-level annotations with semantic categories overlaid on a keyframe. (b) presents cropped word images extracted from those annotations, including handwritten (top) and printed (bottom) examples.}
  \label{fig:clapper_overview}
\end{figure*}

Figure~\ref{fig:clapper_overview} shows examples of annotated frames and cropped text samples. Frame annotations include word-level bounding polygons, with each box representing a single word. Box colors correspond to different semantic categories (Figure~\ref{fig:anno_frame}). In addition, each frame includes a \texttt{Frame Window} annotation indicating the valid region of interest for that frame. Each annotated text instance is exported as an individual cropped image to support word-level recognition tasks. These crops are generated based on the rotated bounding boxes, preserving original text orientation and minimizing background noise. Cropped samples are used in recognition experiments and provide a standardized, text-centered input format for OCR benchmarking. Cropped word images (Figure~\ref{fig:crop_collage}) capture stylistic variability across handwritten and printed scripts. These crops are useful for recognition tasks and style-aware analysis.

Each annotated text instance includes three spatial representations: a rotated bounding box with orientation, a four-point polygon derived from its corners, and an axis-aligned bounding box for compatibility with standard detection models. In addition to spatial information, each instance is annotated with word-level transcription, a semantic category label (\texttt{Text}, \texttt{Date}, \texttt{Location}, \texttt{Recorded\_By}, or \texttt{Attribute}), a handwritten flag indicating whether the text is handwritten or printed, and an occlusion status. The visible content is labeled with \texttt{?} placeholders in the transcription field for partially occluded text. Adjacent frames were used to recover missing characters when possible, and those characters were recorded separately in the \texttt{occluded\_chars} field. If a character remained consistently occluded across all available frames and could not be confidently inferred, it was excluded from the annotation. All recorded occluded characters passed human review to ensure visual plausibility.

\subsection{Dataset Split and Statistical Analysis}\label{sec:dataset_splits}

We structured the ClapperText dataset into train, validation, and test subsets with strict video-level disjointness to support both zero-shot and few-shot learning scenarios. Frames originating from the same video segment are never split across different sets. The training set consists of only 18 video segments, with 8 for validation. The remaining 101 segments form the test set. This split deliberately reflects the low-resource setting faced in historical document analysis, where labeled data is scarce and generalization across domains is critical. Our experiments benchmark multiple models by comparing zero-shot and fine-tuned performance, demonstrating the value of ClapperText for few-shot learning and revealing how even limited supervision can significantly improve OCR accuracy in degraded archival settings.

Table \ref{tab:split_statistics} summarises the statistics of the training, validation, and test splits. For each subset we list the number of videos, frames, and word annotations, together with the percentages of handwritten words and occluded instances. More than two thirds of all annotations are handwritten, and occlusions—though infrequent—occur in every split. The test set alone contains more than 70 000 labeled instances, providing a solid basis for rigorous evaluation across multiple OCR sub-tasks.

\begin{table}[htb]
    \centering
    \caption{ClapperText dataset statistics across different splits. HW and Occ. refer to the percentage of handwritten and occluded annotations, respectively.}
    \label{tab:split_statistics}
    \begin{tabular}{lcccccccc}
        \toprule
         & \multirow{2}{*}{\textbf{Videos}} & \multicolumn{2}{c}{\textbf{All Frames}} & \multicolumn{2}{c}{\textbf{Keyframes}} & \multirow{2}{*}{\textbf{HW (\%)}} & \multirow{2}{*}{\textbf{Occ. (\%)}} \\
        \cmidrule(lr){3-4} \cmidrule(lr){5-6}
        & & Frames & Annotations & Frames & Annotations & & \\
        \midrule
        Train  & 18  & 1122 & 17,749 & 238  & 4,962  & 72.3\% & 4.8\% \\
        Val    & 8   & 527  & 4,983  & 91   & 1,278  & 67.6\% & 1.6\% \\
        Test   & 101 & 8164 & 71,841 & 1531 & 20,947 & 66.2\% & 0.9\% \\
        \textbf{Total} & 127 & 9813 & 94,573 & 1860 & 27,187 & 67.4\% & 1.7\% \\
        \bottomrule
    \end{tabular}
\end{table}

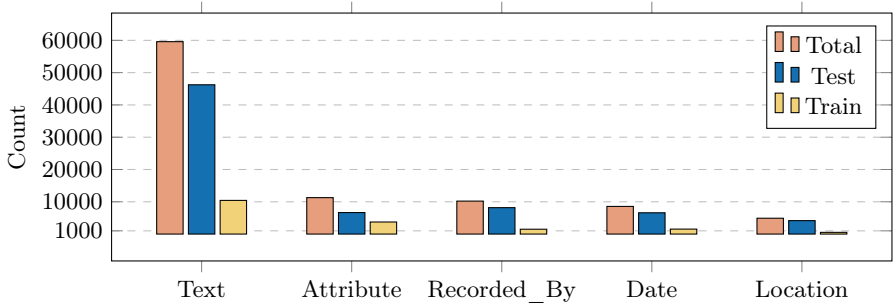
\begin{figure}[htbp]
    \centering


\definecolor{myorange}{RGB}{230,158,125}
\definecolor{myblue}{RGB}{20,112,176}
\definecolor{myyellow}{RGB}{241,210,120}

\begin{tikzpicture}
\begin{axis}[
    ybar,
    width=0.98\textwidth,
    height=0.4\textwidth,
    tick label style={/pgf/number format/fixed},
    scaled ticks=false,
    enlargelimits=0.15,
    ytick={1000,10000, 20000, 30000, 40000, 50000, 60000},
    yticklabels={1000,10000,20000,30000,40000,50000,60000},
    legend style={at={(0.98,0.95)},
      anchor=north east,legend columns=1},
    ylabel={Count},
    symbolic x coords={Text,Attribute,Recorded\_By, Date, Location},
    xtick=data,
    ymajorgrids=true,
    grid style=dashed,
    ]
\addplot[fill=myorange] coordinates {(Text,59595) (Attribute,11286) (Recorded\_By,10213) (Date,8568) (Location,4911)};
\addplot[fill=myblue] coordinates {(Text,46239) (Attribute,6658) (Recorded\_By,8184) (Date,6597) (Location,4163)};
\addplot[fill=myyellow] coordinates {(Text,10431) (Attribute,3739) (Recorded\_By,1504) (Date,1541) (Location,534)};
\legend{Total,Test,Train}
\end{axis}
\end{tikzpicture}
    \caption{Comparison of text category distribution in train and test sets.}
    \label{fig:category_statistics}
\end{figure}

Figure \ref{fig:category_statistics} shows the distribution of the five semantic labels (\texttt{Text}, \texttt{Attribute}, \texttt{Recorded\_By}, \texttt{Date}, and \texttt{Location}). The relative frequencies of these categories are closely matched between the training and test sets, ensuring balanced coverage for model generalisation.

A substantial proportion of annotations are handwritten (67.4\%), adding complexity to detection and recognition tasks. This imbalance reflects real-world archival sources, where handwriting often dominates. Figure~\ref{fig:split_examples} further visualizes the visual diversity across splits by showing representative samples from the train and test sets. Despite the limited training size, the split captures varied writing styles and backgrounds consistent with the test domain. The cropped word-level images (see Figure~\ref{fig:crop_collage}) contain various writing styles, font structures, and pen conditions, making ClapperText a suitable benchmark for handwriting-robust OCR systems.

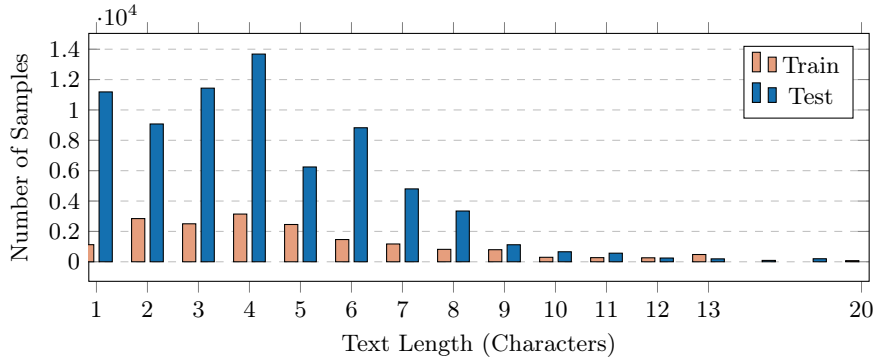
\begin{figure}[htbp]
    \centering
    \definecolor{myorange}{RGB}{230,158,125}
\definecolor{myblue}{RGB}{20,112,176}
\definecolor{myyellow}{RGB}{241,210,120}




\begin{tikzpicture}
\begin{axis}[
    ybar,
    width=0.98\textwidth,
    height=0.4\textwidth,
    bar width=5pt,
    enlarge x limits=0.01,
    ylabel={Number of Samples},
    xlabel={Text Length (Characters)},
    symbolic x coords={1,2,3,4,5,6,7,8,9,10,11,12,13,14,16,20},
    xtick=data,
    ytick={0,2000,4000,6000,8000,10000,12000,14000},
    legend style={at={(0.98,0.95)}, anchor=north east},
    ymajorgrids=true,
    grid style=dashed,
]
\addplot[fill=myorange] coordinates {
    (1,1126) (2,2844) (3,2501) (4,3142) (5,2455) (6,1464) (7,1172)
    (8,818) (9,795) (10,295) (11,272) (12,262) (13,478) (20,73)
};

\addplot[fill=myblue] coordinates {
    (1,11187) (2,9076) (3,11436) (4,13680) (5,6244) (6,8827) (7,4800)
    (8,3343) (9,1121) (10,659) (11,566) (12,247) (13,192) (14,90) (16,200)
};

\legend{Train, Test}
\end{axis}
\end{tikzpicture}
    \caption{Histogram of text length distribution in the train and test sets.}
    \label{fig:text_length_distribution}
\end{figure}

Word length is another factor that influences recognition difficulty. Figure~\ref{fig:text_length_distribution} presents a histogram of character counts per word instance in the train and test splits. Both subsets exhibit a strong concentration of short words, yet the mean word length is slightly higher in the training set (7.91 characters) than in the test set (6.95 characters). This greater lexical variety in the training data is helpful for model fine-tuning.

\section{Benchmarking Experiments}

The benchmarking experiments evaluate the performance of representative text recognition and detection models on the ClapperText dataset. This section presents the experimental setup and results across two primary tasks: text recognition on cropped text images and text detection on full video frames.

\subsection{Experimental Setup}

All experiments were conducted using the MMOCR framework with official pre-trained weights. We evaluated both zero-shot and fine-tuned performance. Models were fine-tuned solely on the ClapperText training split (18 videos), validated on 8 videos, and tested on keyframes from 101 videos reserved for testing. To ensure balanced evaluation, metrics were computed per video and averaged over the test set. During training, each epoch sampled up to 20 frames per video, prioritizing keyframes. For videos with fewer than 20 frames, frames were resampled until 20 selections were obtained. All models were trained for up to 36 epochs with early stopping after 18 epochs of no improvement. Default learning schedules from MMOCR were used.

For text recognition, we assessed six widely used models: CRNN \cite{shi2016end}, MASTER \cite{Lu2021MASTER}, NRTR \cite{sheng2019nrtr}, RobustScanner \cite{yue2020robustscanner}, SAR \cite{li2019show}, and SVTR \cite{ijcai2022p124}. These models represent a variety of architectures, including CNN-based approaches such as CRNN, as well as transformer-based methods such as NRTR. The models were pre-trained on different synthetic and real-world datasets. Word Recognition Accuracy (WRA) is used as the main metric, with normalization applied to casing and symbols. For text detection, we evaluated DBNet \cite{Liao_Wan_Yao_Chen_Bai_2020}, DBNet++ \cite{liao2022real}, FCENet \cite{zhu2021fourier}, Mask R-CNN \cite{8237584}, PANet \cite{WangXSZWLYS19}, PSENet \cite{wang2019shape}, and TextSnake \cite{long2018textsnake}, covering both segmentation-based and region proposal-based detection paradigms. 

\subsection{Text Recognition on Cropped Image Dataset}
\label{sec:recognition_results}

We evaluated six representative text recognition models on the cropped-word subset of ClapperText. All models were tested in both \emph{zero-shot} and \emph{fine-tuned} settings. In the zero-shot setting, we used the official MMOCR pre-trained weights. For fine-tuning, models were trained on the 18-video training split and selected using the 8-video validation split, as described in Section~\ref{sec:dataset_splits}. Table~\ref{tab:rec_main} compares the word recognition accuracy (WRA) of each model across three domains: regular text (IIIT5K, SVT, IC13), irregular text (IC15, SVTP, CUTE), and our ClapperText dataset. As shown, all models experience a substantial drop in zero-shot performance on ClapperText, despite achieving high accuracy on traditional benchmarks. For instance, NRTR-R31 (1/8) achieves over 94\% on regular benchmarks but only 67.46\% on ClapperText, confirming the domain gap between ClapperText and existing datasets.

\begin{table}[htb]
\centering
\small
\caption{Word-accuracy (\%) after case\,+\,symbol normalisation.  
Regular-/Irregular-Text numbers are reproduced from public benchmarks.}
\resizebox{\textwidth}{!}{
\begin{tabular}{lCCCCCCDD}
\toprule
\multirow{2}{*}{\textbf{Model}} &
\multicolumn{3}{c}{\textbf{Regular Text}} &
\multicolumn{3}{c}{\textbf{Irregular Text}} &
\multicolumn{2}{c}{\textbf{ClapperText}} \\
\cmidrule(lr){2-4}\cmidrule(lr){5-7}\cmidrule(lr){8-9}
& IIIT5K & SVT & IC13 & IC15 & SVTP & CUTE & Zero-Shot & Fine-Tuned \\
\midrule
CRNN (Mini-VGG)          & 80.53 & 79.91 & 87.39 & 55.71 & 60.93 & 56.94 & 30.99 & 54.62 \\
MASTER                   & 94.90 & 88.87 & \textbf{95.17} & 76.50 & 84.65 & 88.89 & 67.98 & 72.48 \\
NRTR (Mod-Trans.)        & 91.47 & 88.41 & 93.69 & 72.46 & 77.83 & 75.00 & \textbf{69.57} & 75.16 \\
NRTR-R31 (1/16)          & 94.70 & 89.18 & 93.99 & 73.76 & 79.69 & 88.54 & 65.56 & \textbf{77.24} \\
NRTR-R31 (1/8)           & 94.83 & 89.18 & 95.07 & 75.78 & 80.16 & 88.89 & 67.46 & 72.66 \\
RobustScanner            & 95.10 & 90.11 & 93.20 & 75.78 & 80.78 & 87.50 & 66.27 & 69.34 \\
SAR (Parallel)           & 95.33 & 89.64 & 93.69 & 76.02 & 83.26 & \textbf{90.62} & 66.25 & 71.91 \\
SAR (Sequential)         & \textbf{95.53} & 90.73 & 94.09 & \textbf{77.61} & 80.93 & 89.58 & 65.60 & 73.35 \\
SVTR-Base                & 85.70 & \textbf{91.81} & 94.38 & 74.48 & 83.88 & 90.28 & 55.68 & 57.60 \\
SVTR-Small               & 85.53 & 90.26 & 94.48 & 74.96 & \textbf{84.96} & 88.54 & 56.97 & 60.24 \\
\bottomrule
\end{tabular}}
\label{tab:rec_main}
\end{table}

Fine-tuning on the ClapperText training set leads to consistent improvements across all models. NRTR-R31 (1/16), for example, improves from 65.56\% to 77.24\% after fine-tuning. MASTER, SAR, and RobustScanner also show gains of 5–10 points, highlighting the importance of domain-specific adaptation. These results emphasize the few-shot learning value of ClapperText: Even with only 18 training videos, models are able to significantly bridge the performance gap.

Table~\ref{tab:rec_detail} provides a more fine-grained analysis of performance on ClapperText, breaking results down into handwritten and printed subsets. Consistent with the dataset’s composition (Section~\ref{sec:dataset_splits}), handwritten samples pose a significantly greater challenge in the zero-shot setting, with models like CRNN and SAR performing 10–15 points lower on handwritten text than printed. NRTR (Mod-Trans.) achieves the highest zero-shot accuracy for both handwritten (63.35\%) and printed text (81.04\%). Fine-tuning closes this gap: nearly all models exhibit greater gains on handwritten text than printed text, indicating that domain adaptation benefits handwritten OCR. MASTER and SAR (Sequential), for example, improve by more than 8 points on handwritten text after fine-tuning.

Partially occluded samples are excluded from these comparisons, as their recognition differs from conventional text recognition tasks. For reference, NRTR (Mod-Trans.) achieves 18.06\% accuracy on occluded words in the zero-shot setting and 30.14\% after fine-tuning, highlighting the difficulty of occlusion and the potential of ClapperText for future research.

\begin{table}[htb]
\centering
\small
\caption{Word-accuracy (\%) on ClapperText after case\,+\,symbol normalisation.  
All metrics are computed on non–occluded samples only.}
\resizebox{\textwidth}{!}{
\begin{tabular}{lcccccc}
\toprule
\multirow{2}{*}{\textbf{Model}} &
\multicolumn{2}{c}{\textbf{All}} &
\multicolumn{2}{c}{\textbf{Handwritten}} &
\multicolumn{2}{c}{\textbf{Printed}} \\
\cmidrule(lr){2-3}\cmidrule(lr){4-5}\cmidrule(lr){6-7}
& Zero-Shot & Fine-Tuned & Zero-Shot & Fine-Tuned & Zero-Shot & Fine-Tuned \\
\midrule
CRNN (Mini-VGG)          & 30.99 & 54.62 & 25.48 & 46.34 & 39.84 & 74.56 \\
MASTER                   & 67.98 & 72.48 & 62.53 & 67.26 & 75.20 & 81.54 \\
NRTR (Mod-Trans.)        & \textbf{69.57} & 75.16 & \textbf{63.35} & 70.68 & \textbf{81.04} & \textbf{84.08} \\
NRTR-R31 (1/16)          & 65.56 & \textbf{77.24} & 59.31 & \textbf{72.62} & 78.04 & 83.96 \\
NRTR-R31 (1/8)           & 67.46 & 72.66 & 61.64 & 68.78 & 77.21 & 80.05 \\
RobustScanner            & 66.27 & 69.34 & 59.74 & 63.60 & 76.10 & 81.18 \\
SAR (Parallel)           & 66.25 & 71.91 & 59.61 & 66.51 & 79.06 & 82.67 \\
SAR (Sequential)         & 65.60 & 73.35 & 59.19 & 67.96 & 76.04 & 83.16 \\
SVTR-Base                & 55.68 & 57.60 & 51.07 & 53.70 & 59.82 & 63.65 \\
SVTR-Small               & 56.97 & 60.24 & 52.36 & 56.63 & 60.96 & 65.73 \\
\bottomrule
\end{tabular}}
\label{tab:rec_detail}
\end{table}

\begin{figure}[htb]
    \centering
    \includegraphics[width=\linewidth]{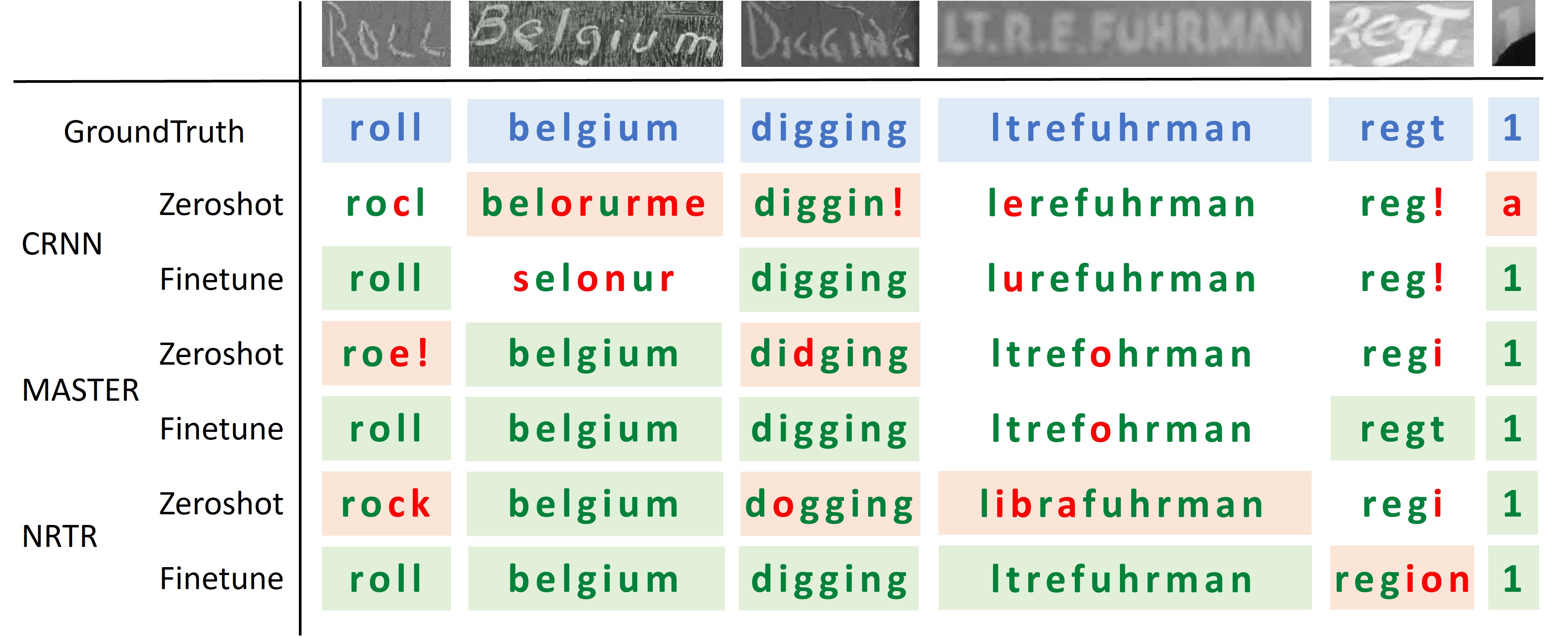}
    \caption{
    Recognition outputs on five ClapperText samples by CRNN, MASTER, and NRTR (zero-shot and fine-tuned). Correct characters are green, errors red; background shading marks best (green) and worst (red) predictions. \texttt{!} indicates missing characters.}
    \label{fig:rec_qualitative}
\end{figure}

\paragraph{Qualitative analysis.} Figure~\ref{fig:rec_qualitative} presents qualitative examples comparing zero-shot and fine-tuned predictions on selected word images to illustrate the recognition characteristics of different models better. Each column corresponds to a different cropped word from ClapperText, chosen to reflect diverse challenges in handwriting style, semantic familiarity, and visual distortion.

In the first column, all models fail to recognize the word \texttt{roll} in the zero-shot setting, despite it being a familiar English word. This is likely due to the unusual handwriting style; fine-tuning corrects this error across all models. The second column, \texttt{belgium}, shows a semantically clear and lexically distinct word. Its unique character sequence likely aids language-aware models like MASTER and NRTR in correctly predicting it even without fine-tuning, whereas CRNN—lacking strong contextual modeling—fails in both settings. The third column exhibits a similar trend to the first: a structurally straightforward word, \texttt{digging}, is initially misrecognized by most models, but fine-tuning resolves the issue.

The fourth column, \texttt{ltrefuhrman}, is a visually clean yet semantically unfamiliar surname. CRNN and MASTER still fail after fine-tuning, whereas NRTR successfully decodes it, demonstrating strong generalization to visually complex words that lack strong language priors. In the fifth column, all models struggle to correctly recognize the final handwritten character \texttt{T}. In contrast, the more structured \texttt{T} in the previous column poses less difficulty. Fine-tuned NRTR mispredicts the word \texttt{regt} as a more semantically plausible alternative, revealing a limitation of its language-driven bias when handling abbreviations or non-lexical tokens. Finally, the last column shows a partially occluded digit \texttt{1}. While CRNN misclassifies it in the zero-shot setting, it is correctly recognized after fine-tuning. This confirms that exposure to even modest occlusion during training improves recognition robustness.

Overall, the qualitative results reinforce the quantitative findings: fine-tuning significantly improves recognition accuracy, especially for handwritten and occluded samples. NRTR consistently demonstrates stronger adaptation to ClapperText’s visual and semantic peculiarities.

\paragraph{Ablation study.} Table~\ref{tab:nrtr_ablation} shows the impact of removing each augmentation module: geometric transformations, random rescaling, and color jittering. 

\begin{table}[htb]
\centering
\small
\caption{Data-augmentation ablation on NRTR-R31 (1/16).  Accuracy reported on validation set.  (\checkmark enabled)}
\begin{tabular}{lccc|c}
\multicolumn{5}{c}{\textbf{A. Global Ablation: turning modules off}}\\\hline
Experiment & Geom & Rescale & ColorJitter & Acc (\%)\\\hline
All Augmentations        & \checkmark & \checkmark & \checkmark & \textbf{68.44} \\
No Geom                  &    & \checkmark & \checkmark & 66.84 \\
No Rescale               & \checkmark &    & \checkmark & 68.18 \\
No ColorJitter           & \checkmark & \checkmark &    & 67.51 \\
No Augmentation (Minimal)&    &    &    & 66.18 \\\hline
\end{tabular}
\label{tab:nrtr_ablation}
\end{table}

To guide the fine-tuning process, we conducted a targeted ablation study on the NRTR-R31 (1/16) model using the training and validation splits. While each module contributes to performance, geometric transformations appear most critical, as disabling them leads to a 1.6-point drop in accuracy. Rescaling also plays a complementary role, especially when combined with other augmentations. Removing all augmentations results in a significant drop of over 2 points compared to the full setup. Based on these results, we enabled all three augmentation types for all fine-tuned models.

\subsection{Text Detection on Video Frames}
\label{sec:detection_results}

Matching between predicted and ground-truth boxes was conducted using a best-match assignment strategy based on polygon IoU, with a threshold of 0.5. Hmean is reported for each video, then averaged across the test set. All detection models were evaluated at their recommended input sizes. Table~\ref{tab:det_main} presents the detection results.

\begin{table}[htb]
\centering
\small
\caption{Detection performance on ClapperText.  
All numbers are Hmean (\%) at IoU\,0.5. Best scores per pre-training dataset are in \textbf{bold}. Deployment metrics (batch\,=\,1).}
\resizebox{\textwidth}{!}{
\begin{tabular}{lcccccc}
\toprule
\textbf{Model} & \textbf{Pretrain} & \multicolumn{2}{c}{\textbf{ClapperText}} & \textbf{FPS} & \textbf{Mem (MB)} \\
\cmidrule(lr){3-4}
& & \textbf{Zero-Shot} & \textbf{Fine-Tuned} & & \\
\midrule
\multicolumn{6}{l}{\textbf{Pre-trained on IC15}} \\

\arrayrulecolor{gray}\midrule[0.3pt]\arrayrulecolor{black}
PANet R18                       & 78.48 & 31.05 & 64.96 & 117.7 & 283 \\
DBNet R18                       & 81.69 & 47.55 & 65.44 & 132.9 & 304 \\
DBNet R50                       & 85.04 & 40.63 & 61.13 & 12.7  & 1542 \\
DBNet R50 + DCN                 & 85.43 & 47.27 & 64.36 & 11.6  & 1545 \\
DBNet R50 + OCLIP               & 86.44 & 40.86 & 62.04 & 12.1  & 1551 \\
DBNet++ R50                     & 86.22 & 38.51 & 57.67 & 10.8  & 2346 \\
DBNet++ R50 + DCN               & 86.84 & \textbf{59.48} & \textbf{68.42} & 9.5   & 2350 \\
DBNet++ R50 + OCLIP             & \textbf{88.82} & 45.73 & 59.81 & 9.8   & 2360 \\
Mask R-CNN R50                  & 81.82 & 17.38 & 58.59 & 7.8   & 1137 \\
Mask R-CNN R50 + OCLIP          & 85.13 & 23.03 & 58.97 & 7.4   & 1158 \\
PSENet R50                      & 77.93 & 17.35 & 30.25 & 6.4   & 4966 \\
PSENet R50 + OCLIP              & 80.37 & 21.14 & 27.97 & 6.3   & 4970 \\
FCENet R50                      & 85.28 & 28.57 & 40.37 & 11.6  & 1232 \\
FCENet R50 + OCLIP              & 86.04 & 22.55 & 43.69 & 10.7  & 1244 \\

\arrayrulecolor{gray}\midrule[0.3pt]\arrayrulecolor{black}
\multicolumn{6}{l}{\textbf{Pre-trained on CTW}} \\
\arrayrulecolor{gray}\midrule[0.3pt]\arrayrulecolor{black}

FCENet R50 + DCN                & 84.88 & 29.07 & 49.99 & 9.4   & 1239 \\
FCENet R50 + OCLIP              & 81.92 & 26.89 & 43.97 & 10.6  & 1244 \\
TextSnake R50                   & 82.86 & 32.70 & 67.29 & 40.5  & 855 \\
TextSnake R50 + OCLIP           & \textbf{85.29} & \textbf{35.41} & \textbf{69.63} & 36.4  & 736 \\
\bottomrule
\end{tabular}}
\label{tab:det_main}
\end{table}

As with text recognition, all models exhibit a sharp drop in zero-shot performance when transferred to ClapperText. For instance, DBNet++ (R50+DCN) achieves 86.84\% Hmean on ICDAR 2015 but only 59.48\% zero-shot on ClapperText. Fine-tuning consistently improves detection, with top-performing models such as DBNet++ (R50+DCN) and TextSnake (R50+OCLIP) achieving 68.42\% and 69.63\% Hmean, respectively. In addition to accuracy, inference speed is also an important consideration for text detection in video. While DBNet++ (R50+DCN) and TextSnake (R50+OCLIP) achieve comparable detection performance, the latter demonstrates substantially higher inference speed under batch size 1, reaching 36.4 FPS compared to 9.5 FPS for DBNet++ (R50+DCN). This throughput is sufficient for real-time processing of 24 FPS archival video. Table~\ref{tab:det_main} reports deployment metrics for all evaluated models, providing practical insights into their suitability for real-world applications.

Beyond the baseline comparisons, we further evaluated how architectural choices and pre-training strategies affect transfer performance. Table~\ref{tab:det_main} reports results for models incorporating Deformable Convolutional Networks (DCN) and OCLIP, a CLIP-based visual encoder enhancement. After fine-tuning on ClapperText, DCN consistently outperforms OCLIP. For example, DBNet++ (R50+DCN) achieves 68.42\% Hmean, compared to 59.81\% for its OCLIP variant. This contrasts with results on ICDAR 2015, where OCLIP provides larger gains. Both variants still improve over the vanilla backbone. A possible explanation is that OCLIP, pre-trained on diverse modern scene-text corpora, is less effective when domain-specific cues such as clapperboard layout are not represented during pre-training. We also examined the influence of pre-training datasets. FCENet, for instance, was initialized using either ICDAR 2015 or CTW1500. Its zero-shot performance on ClapperText improves from 22.55\% with ICDAR pre-training to 26.89\% with CTW. After fine-tuning, however, both versions reach similar accuracy (43.69\% and 43.97\%, respectively). This indicates that fine-tuning reduces the impact of pre-training differences, although better initialization can still provide advantages in low-resource scenarios.

\begin{figure}[htb]
    \centering
    \includegraphics[width=\linewidth]{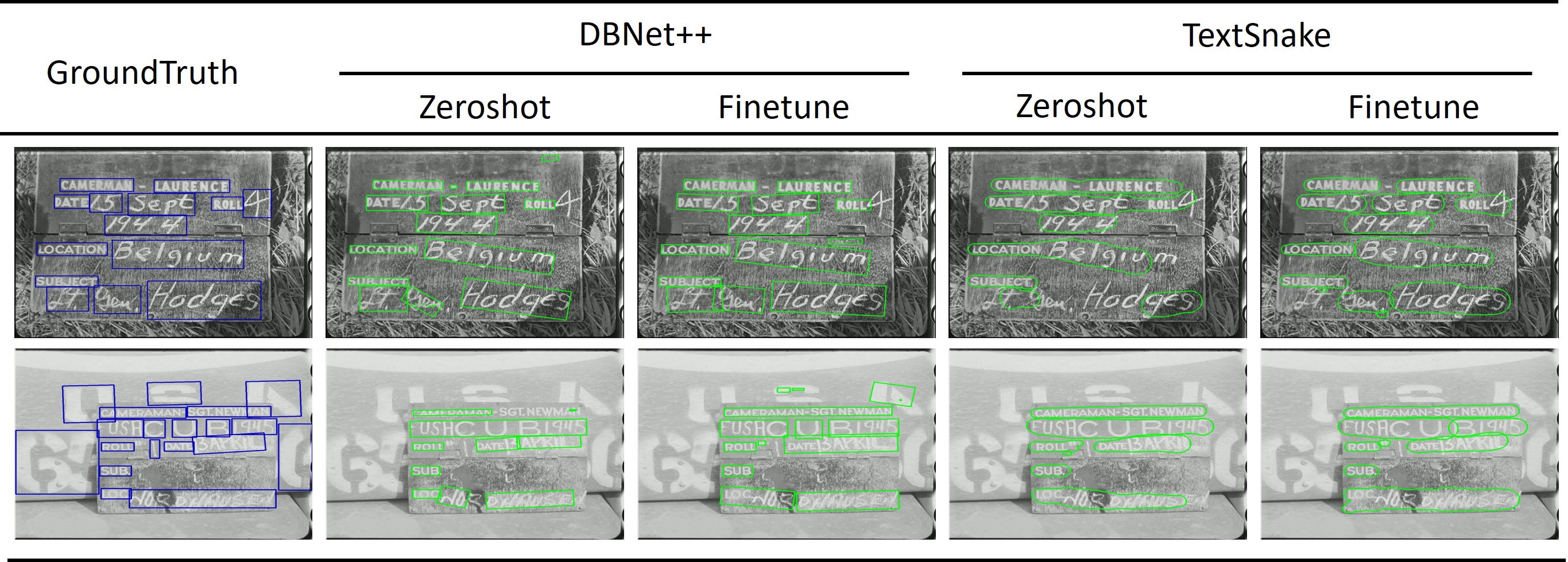}
    \caption{Detection outputs from DBNet++ (R50+DCN) and TextSnake (R50+OCLIP) in zero-shot and fine-tuned settings. Top: structured layout with handwriting; bottom: cluttered background. Fine-tuning improves alignment, though over-grouping (TextSnake) and false positives (DBNet++) remain.}

    \label{fig:det_vis}
\end{figure}

\paragraph{Qualitative analysis.}
Figure~\ref{fig:det_vis} compares the detection outputs of DBNet++ (R50+DCN) and TextSnake (R50+OCLIP) under zero-shot and fine-tuned settings on two challenging ClapperText frames. While both models benefit from fine-tuning, they exhibit different behaviors regarding boundary precision and structural interpretation.

In the first row, the clapperboard contains well-aligned fields, but also includes handwritten words with irregular spacing and shape. TextSnake in the zero-shot setting tends to over-group adjacent words on the same line, treating multi-word fields as a single region. After fine-tuning, it better separates words, but fails to locate boundaries accurately, notably when the word lacks strong semantic priors or exhibits unusual handwriting. In contrast, DBNet++ achieves reasonable localization even in the zero-shot condition and further refines box tightness after fine-tuning, although some words remain undetected.

The second row presents a more difficult case with substantial background text interference. The clapperboard is partially obstructed and overlaps with military stenciling in the background. DBNet++ shows signs of confusion after fine-tuning, mistakenly detecting background text near the lower edge. TextSnake avoids this issue but also misses smaller clapperboard fields. These examples illustrate how fine-tuning improves both models, but also expose their respective limitations: DBNet++ is more layout-sensitive, while TextSnake tends to smooth over structural distinctions.

\paragraph{Ablation study.} To identify effective augmentation strategies for historical document detection, we performed ablation experiments on DBNet++ (ResNet-50). Results show that random cropping and multi-scale resizing are the most impactful augmentations, boosting Hmean from 65.82\% (no augmentation) to 72.45\%. Removing cropping or scaling led to significant drops in accuracy (–5.8 and –0.3 points, respectively), while color jitter had minimal effect. Among individual augmentations, scale alone achieved the highest Hmean (72.45\%), confirming its robustness across varied frame conditions. These findings guided the final augmentation choices used in model training.

\section{Conclusion}

We introduced ClapperText, a curated dataset of historical clapperboard frames with word-level annotations for both printed and handwritten text under real-world degradation. Systematic benchmarking of modern OCR models demonstrates persistent limitations in handling handwriting, occlusion, and visual noise in archival material. Although fine-tuning yields consistent improvements, challenges remain in recognizing degraded handwriting and resolving partial occlusions under limited supervision. ClapperText provides a valuable testbed for evaluating OCR in low-resource settings and exposes open problems in adapting text models to non-traditional domains. Future research may investigate domain adaptation, temporal modeling, and semantic context integration across frames.

\newpage

\begin{credits}
\subsubsection*{Acknowledgements}
This work was supported by the Austrian Science Fund (FWF) -- doc.funds.connect, under project grant no. DFH 37-N: "Visual Heritage: Visual Analytics and Computer Vision Meet Cultural Heritage.". 
\end{credits}

\bibliographystyle{splncs04}
\bibliography{references}

\end{document}